%% file: main.tex
\documentclass[final]{cvpr}

\usepackage{times}
\usepackage{mathptmx}
\usepackage{mathrsfs}
\usepackage{epsfig}
\usepackage{graphicx}
\usepackage{amsmath}
\usepackage{amssymb}
\usepackage{booktabs}
\usepackage{siunitx}
\usepackage{multirow}
\usepackage{mathtools}
\usepackage{ifthen}
\usepackage{anyfontsize}

\DeclareMathOperator*{\argmax}{arg\,max}

\usepackage[pagebackref=true,breaklinks=true,colorlinks,bookmarks=false]{hyperref}

\begin{document}
\input{macros}

\newcommand{\paper}{main} 
\newcommand{\arxiv}{true} 

\ifthenelse{\equal{\paper}{main}}{
\title{ArtEmis: Affective Language for Visual Art}
}

\ifthenelse{\equal{\paper}{supp}}{
\title{ArtEmis: Affective Language for Visual Art\\Supplemental Material}
}

\author{
\begin{tabular}[t]{c@{\extracolsep{2.0em}}c@{\extracolsep{2.0em}}c@{\extracolsep{2.0em}}c}    
    \multicolumn{3}{c}{    
        \begin{tabular}{c}
        Panos Achlioptas${}^{1}$ \\
        \fontsize{9}{9} \selectfont \texttt{panos@cs.stanford.edu}
        \end{tabular}
        \begin{tabular}{c}
         Maks Ovsjanikov${}^{2}$ \\
         \fontsize{9}{9} \selectfont \texttt{maks@lix.polytechnique.fr}
        \end{tabular}
        \begin{tabular}{c}
         Kilichbek Haydarov${}^{3}$ \\
         \fontsize{9}{9}\selectfont \texttt{kilichbek.haydarov@kaust.edu.sa}
        \end{tabular}}\vspace{3pt}\\
    \multicolumn{3}{c}{
        \begin{tabular}{c}
        Mohamed Elhoseiny${}^{3,1}$ \\
        \fontsize{9}{9}\selectfont \texttt{mohamed.elhoseiny@kaust.edu.sa}
        \end{tabular}
        \begin{tabular}{c}
        Leonidas Guibas${}^{1}$ \\
        \fontsize{9}{9} \selectfont \texttt{guibas@cs.stanford.edu}
        \end{tabular}}\vspace{9pt}\\
    \multicolumn{3}{c}{${}^{1}$Stanford University}\\
    \multicolumn{3}{c}{${}^{2}$LIX, Ecole Polytechnique, IP Paris}\\
    \multicolumn{3}{c}{${}^{3}$King Abdullah University of Science and Technology (KAUST)}\\
\end{tabular}
}

\maketitle

\ifthenelse{\equal{\paper}{main}}{

\input{sections/abstract}

\input{sections/introduction}

\input{sections/related_work}

\input{sections/dataset}

\input{sections/method}
\input{sections/evaluation}
\input{sections/results}
\input{sections/conclusion}
\input{sections/acknowledgments}
}

\ifthenelse{\equal{\paper}{supp}}{
\appendix
\input{sections/supplemental_material}
}

{\small
\bibliographystyle{ieee_fullname}
\bibliography{references}
}

\end{document}

%% file: macros.tex
\newcommand{\E}{\mathbb{E}}  
\def\reals{{\mathbb R}}      

\def\vx{{\bm{x}}}
\def\vy{{\bm{y}}}
\def\vz{{\bm{z}}}
\def\vtheta{{\bm{\theta}}}

\newcommand{\textarrowright}{ $\,\to\,$ }

\newcommand{\tocheck}[1]{{\textcolor{red}{#1}}}
\newcommand{\red}[1]{{\color{red}{#1}}}
\newcommand{\yellow}[1]{{\color{yellow}{#1}}}
\newcommand{\blue}[1]{{\color{blue}{#1}}}
\newcommand{\outline}[1]{}
\newcommand{\fix}{\marginpar{FIX}}
\renewcommand{\tilde}{{\raise.17ex\hbox{$\scriptstyle\sim$}}}
\newcommand{\qq}[2]{\red{{#1}}\blue{#2}}

\newcommand{\coco}{COCO}
\newcommand{\cococite}{\cite{lin14eccv,coco_chen2015}}
\newcommand{\oid}{Open~Images}
\newcommand{\flickr}{Flickr30k Ent.}
\newcommand{\ade}{ADE20K}
\newcommand{\vg}{Visual Genome}

\newcommand{\numcaptions}{439,121}
\newcommand{\numimages}{81,446}
\newcommand{\numimagesshort}{81K}
\newcommand{\numwords}{36,347}
\newcommand{\numworkers}{6,377}
\newcommand{\numhours}{10,220}
\newcommand{\numcaptionsshort}{\num{439}K}
\newcommand{\captionlength}{\num{15.8}}
\newcommand{\numsomethingelse}{49,663}
\newcommand{\datasetname}{ArtEmis}
\newcommand{\suppmat}{Supp.~Mat.}
\newcommand{\webpage}{https://artemisdataset.org}
\newcommand{\numtestutterances}{40,137}
\newcommand{\resnetD}{32}

\newcommand{\mypara}[1]{\vspace*{-8pt}\paragraph{#1}}
\newcommand{\nospacepara}[1]{\noindent{\bf {#1}}}

%% file: sections/abstract.tex

\begin{abstract}
We present a novel large-scale dataset and accompanying machine learning models aimed at providing a detailed understanding of the interplay between visual content, its emotional effect, and explanations for the latter in language. In contrast to most existing annotation datasets in computer vision, we focus on the affective experience triggered by visual artworks and ask the annotators to indicate the dominant emotion they feel for a given image and, crucially, to also provide a grounded verbal explanation for their emotion choice. As we demonstrate below, this leads to a rich set of signals for both the objective content and the affective impact of an image, creating associations with abstract concepts (e.g., ``freedom'' or ``love''), or references that go beyond what is directly visible, including visual similes and metaphors, or subjective references to personal experiences. We focus on visual art (e.g., paintings, artistic photographs) as it is a prime example of imagery created to elicit emotional responses from its viewers. Our dataset, termed \textit{\datasetname}, contains  \numcaptionsshort~emotion attributions and explanations from humans, on \numimagesshort~artworks from WikiArt. Building on this data, we train and demonstrate a series of captioning systems capable of expressing and explaining emotions from visual stimuli. Remarkably, the captions produced by these systems often succeed in reflecting the semantic and abstract content of the image, going well beyond systems trained on existing datasets. The collected dataset and developed methods are available at \url{\webpage}.
\end{abstract}

%% file: sections/introduction.tex

\begin{figure*}[ht]
  \includegraphics[width=\linewidth]{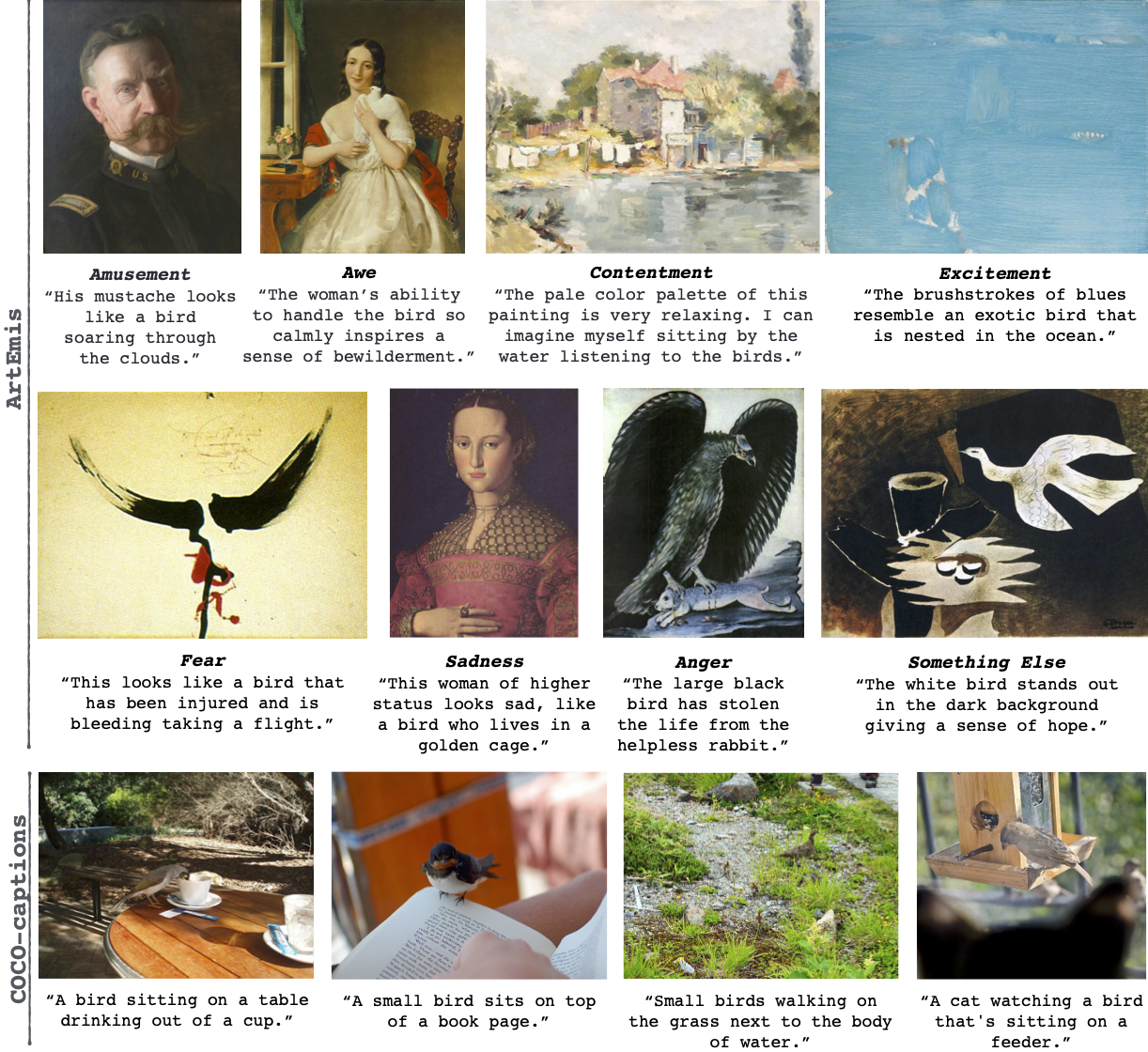}    
  \caption{{\bf Examples of affective explanations vs.~content-based captions mentioning the word `bird'.} The content-based annotations are from COCO-captions~\cite{coco_chen2015} (bottom row), where each utterance refers to objects and actions directly visible in each corresponding image. In ArtEmis (top and middle rows) the annotators expose a wide range of abstract semantics and emotional states associated with the concept of a bird when attempting to explain their primary emotion (shown in boldface). The exposed semantics include properties that are not directly visible: \textit{birds can be listened to, they fly, they can bring hope, but also can be sad when they are in `golden cages'.}}
  \label{fig:bird_annotations}
\end{figure*}

\section{Introduction}
\label{sec:introduction}
Emotions are among the most pervasive aspects of human experience. While emotions are not themselves linguistic constructs, 
the most robust and permanent access we have to them is through language~\cite{OrtonyBook}. In this work, we focus on collecting and analyzing at scale language that \textit{explains} emotions generated by observing visual artworks. Specifically, we seek to better understand the link between the visual properties of an artwork, the possibly subjective affective experience that it produces, and the way such emotions are explained via language. Building on this data and recent machine learning approaches, we also design and test neural-based speakers that aim to emulate human emotional responses to visual art and provide associated explanations.
\mypara{Why visual art?}
We focus on visual artworks for two reasons. First and foremost because art is often created with the intent of provoking emotional reactions from its viewers. In the words of Leo Tolstoy,\textit{``art is a human activity consisting in that one human consciously hands on to others feelings they have lived through, and that other people are infected by these feelings, and also experience them''}~\cite{tolstoy-art}. Second, artworks, and abstract forms of art in particular, often defy simple explanations and might not have a single, easily-identifiable subject or label. Therefore, an affective response may require a more detailed analysis integrating the image content as well as its effect on the viewer. This is unlike most natural images that are commonly labeled through purely objective content-based labeling mechanisms based on the objects or actions they include \cite{coco_chen2015,caba2015activitynet}. Instead, by focusing on art, we aim to initiate a more nuanced perceptual image understanding which, downstream, can also be applied to richer understanding of ordinary images. 

We begin this effort by introducing a large-scale dataset termed \textit{\datasetname} [\underline{Art} \underline{Em}ot\underline{i}on\underline{s}] that associates human emotions with artworks and contains \textit{explanations} in natural language of the rationale behind each triggered emotion.

\mypara{Novelty of  \datasetname.}
Our dataset is novel as it concerns an underexplored problem in computer vision: the formation of linguistic affective explanations grounded on visual stimuli. Specifically, \datasetname~ exposes moods, feelings, personal attitudes, but also abstract concepts like freedom or love, grounded over a wide variety of complex visual stimuli (see Section~\ref{para:emotion-analysis}). The annotators typically explain and link visual attributes to psychological interpretations e.g., \textit{`her youthful face accentuates her innocence'}, highlight peculiarities of displayed subjects, e.g., \textit{`her neck is too long, this seems unnatural'}; and include imaginative or metaphorical descriptions of objects that do not directly appear in the image but may relate to the subject's experience; \textit{`it reminds me of my grandmother' or `it looks like blood'} (over $20\%$ of our corpus contains such similes).

\mypara{Subjectivity of responses.}
Unlike existing captioning datasets, \datasetname~welcomes the subjective and personal angle that an emotional explanation (in the form of a caption) might have. Even a single person can have a range of emotional reactions to a given stimulus \cite{Mikels_2005,compare_dim_models,vector_model_of_affect,circumplex_model} and, as shown in  Fig.~\ref{fig:subjectivity-of-artemis}, this is amplified across different annotators. The subjectivity and rich semantic content  distinguish {\datasetname} from, e.g., the widely used COCO dataset~\cite{coco_chen2015}. Fig.~\ref{fig:bird_annotations} shows different images from both {\datasetname} and COCO datasets with captions including the word \textit{bird}, where the imaginative and metaphorical nature of {\datasetname} is apparent (e.g., `bird gives hope' and `life as a caged bird'). Interestingly, despite this phenomenon, as we show later (Section~\ref{para:emotion-analysis}), (1) there is often substantial agreement among annotators regarding their \textit{dominant} emotional reactions, and (2) our collected explanations are often \textit{pragmatic} -- i.e., they also contain  references to visual elements present in the image (see Section \ref{para:dataset-Maturity-reasonableness-specificity}).

\mypara{Difficulty of emotional explanations.}
There is debate within the neuroscience community on whether human emotions are innate, generated by patterns of neural activity, or learned \cite{shackman2019emotional,adolphs2017should,barrett2019historical}. There may be intrinsic difficulties with producing emotion explanations in language -- thus the task can be challenging for annotators in ways that traditional image captioning is not. Our approach is supported by significant research that argues for the central role of language in capturing and even helping to form emotions \cite{lindquist2015role,barrett2017emotions}, including the \emph{Theory of Constructed Emotions}~\cite{barrett2017emotions,barrett2017theory,barrett2006solving,barrett2007mice} by Lisa Feldman Barrett. Nevertheless, this debate suggests that caution is needed when comparing, under various standard metrics, {\datasetname} with other captioning datasets.

\mypara{Affective neural speakers.} To further demonstrate the potential of {\datasetname}, we experimented with building a number of neural speakers, using deep learning language generation techniques trained on our dataset. The best of our speakers often produce well-grounded affective explanations, respond to abstract visual stimuli, and fare reasonably well in emotional Turing tests, even when competing with humans.

\noindent In summary, we make the following key contributions:
\begin{itemize}
	\item We introduce \textit{\datasetname}, a large scale dataset of emotional reactions to visual artwork coupled with explanations of these emotions in language (Section~\ref{sec:dataset}).
	\item We show how the collected corpus contains utterances that are significantly more affective, abstract, and rich with metaphors and similes, compared to existing datasets (Sections \ref{para:linguistic_analysis}-\ref{para:emotion-analysis}). 	
	\item Using \textit{\datasetname}, we develop machine learning models for dominant emotion prediction from images or text, and neural speakers that can produce plausible grounded emotion explanations (Sections \ref{sec:method} and~\ref{sec:experimental_results}).
\end{itemize}

%% file: sections/related_work.tex

\begin{figure}[t!]
  \includegraphics[width=\linewidth]{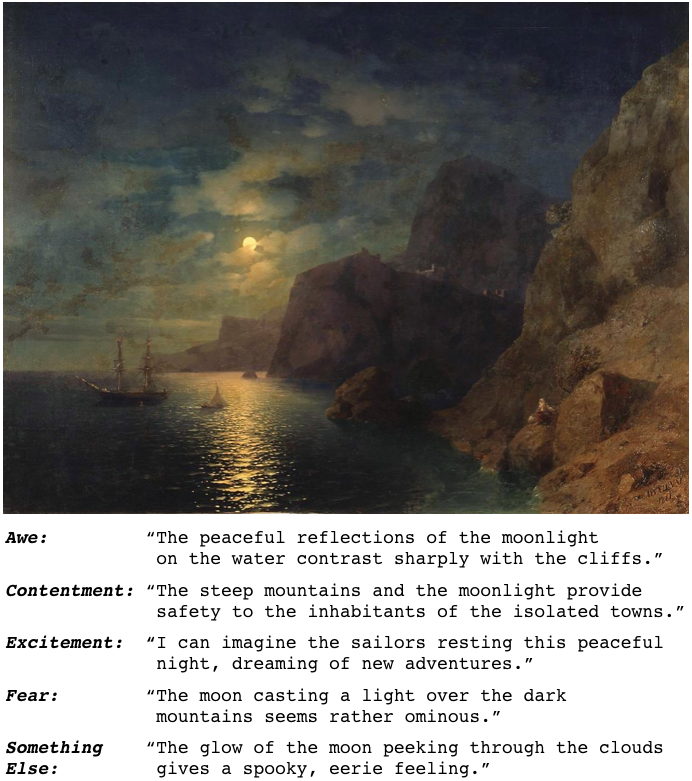}  
  \vspace{3pt}
  \caption{\textbf{Examples of different emotional reactions for the same stimulus.} The emotions experienced (in bold font) for the shown painting vary across annotators \textit{and} are reasonably justified (next to each emotion, the annotator's explanation is given). We note that 61\% of all annotated artworks have at least one positive \textit{and} one negative emotional reaction. See Section~\ref{para:emotion-analysis} for details.}
  \label{fig:subjectivity-of-artemis}
\end{figure}

\section{Background and related work}
\label{sec:related_work}

\paragraph{Emotion classification.}
\label{para:related-work:emotions}
Following previous studies \cite{img_clf_art,Yanulevskaya-emotions,Zhao2014art,emotion_clf}, we adopt throughout this work the same discrete set of eight \textit{categorical} emotion states. Concretely, we consider: \textit{anger}, \textit{disgust}, \textit{fear}, and \textit{sadness} as negative emotions, and \textit{amusement}, \textit{awe}, \textit{contentment}, and \textit{excitement} as positive emotions. The four negative emotions are considered universal and basic (as proposed by Ekman in \cite{ekman_emotions}) and have been shown to capture well the discrete emotions of the International Affective Picture System~\cite{IAPS}. The four positive emotions are finer grained versions of \textit{happiness}~\cite{happiness_1}. We note that while \textit{awe} can be associated with a negative state, following previous works (\cite{Mikels_2005, emotion_clf}), we treat \textit{awe} as a positive emotion in our analyses.

\vspace{-4pt}
\mypara{Deep learning, emotions, and art.}
Most existing works in Computer Vision treat emotions as an image classification problem, and build systems that try to deduce the main/dominant emotion a given image will elicit~\cite{img_clf_art,Yanulevskaya-emotions,Zhao2014art,emotion_clf_valence_arousal}. An interesting work linking paintings to textual descriptions of their historical and social intricacies is given in \cite{sem_art}. Also, the work of ~\cite{prose_for_painting} attempts to make captions for paintings in the prose of Shakespeare using language style transfer. Last, the work of~\cite{Wilber2017BAMTB} introduces a large scale dataset of artistic imagery with multiple attribute annotations. Unlike these works, we focus on developing machine learning tools for analyzing and generating \emph{explanations} of emotions as evoked by artworks.

\mypara{Captioning models and data.} 
There is a lot of work and corresponding captioning datasets~\cite{young14tacl,Kazemzadeh,conceptual-captions,VG_Krishna_2017,mao16,pont2020connecting} that focus on different aspects of human cognition. For instance COCO-captions~\cite{coco_chen2015} concern descriptions of common objects in natural images, the data of Monroe et al.~\cite{monroe_colors} include discriminative references for 2D monochromatic colors, Achlioptas et al. \cite{achlioptas2020referit3d,achlioptas2019shapeglot} collects discriminative utterances for 3D objects, etc. There is correspondingly also a large volume on deep-net based captioning approaches \cite{baby_talk,mao16,unsup_context_aware,licheng_16,nagaraja16,mattnet,nagaraja16}. The seminal works of~\cite{show-tell,karpathy2015deep} opened this path by capitalizing on advancements done in deep recurrent networks (LSTMs~\cite{lstm}), along with other classic ideas like training with Teacher Forcing~\cite{teacher_forcing}. Our neural speakers build on these `standard' techniques, and {\datasetname} adds a new dimension to image-based captioning reflecting emotions.

\mypara{Sentiment-driven captions.}
There exists significantly less captioning work concerning sentiments (positive vs.~negative emotions). Radford and colleagues \cite{unsup_sentiment_cell} discovered that a single unit in recurrent language models trained without sentiment labels, is automatically learning concepts of sentiment; and enables sentiment-oriented manipulation by fixing the sign of that unit. Other early work like SentiCap~\cite{SentiCap} and follow-ups like~\cite{sentiment_inject}, provided explicit sentiment-based supervision to enable sentiment-flavored language generation grounded on real-world images. These studies focus on the visual cues that are responsible for only two emotional reactions (positive and negative) and, most importantly, they do not produce emotion-\emph{explaining} language.

%% file: sections/dataset.tex
\begin{figure}[ht]
  \includegraphics[scale=0.35]{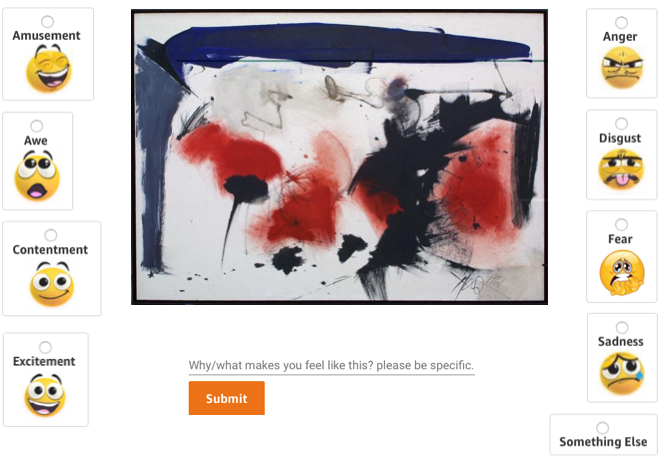}
  \caption{{\bf AMT interface for {\datasetname} data collection.} To ease the cognitive task of self-identifying and correctly selecting the dominant emotion felt by each annotator, we display expressive emojis to accentuate the semantics of the available options.}
  \label{fig:data_collection_interface}
\end{figure}

\begin{figure*}[ht]
  \includegraphics[width=\textwidth]{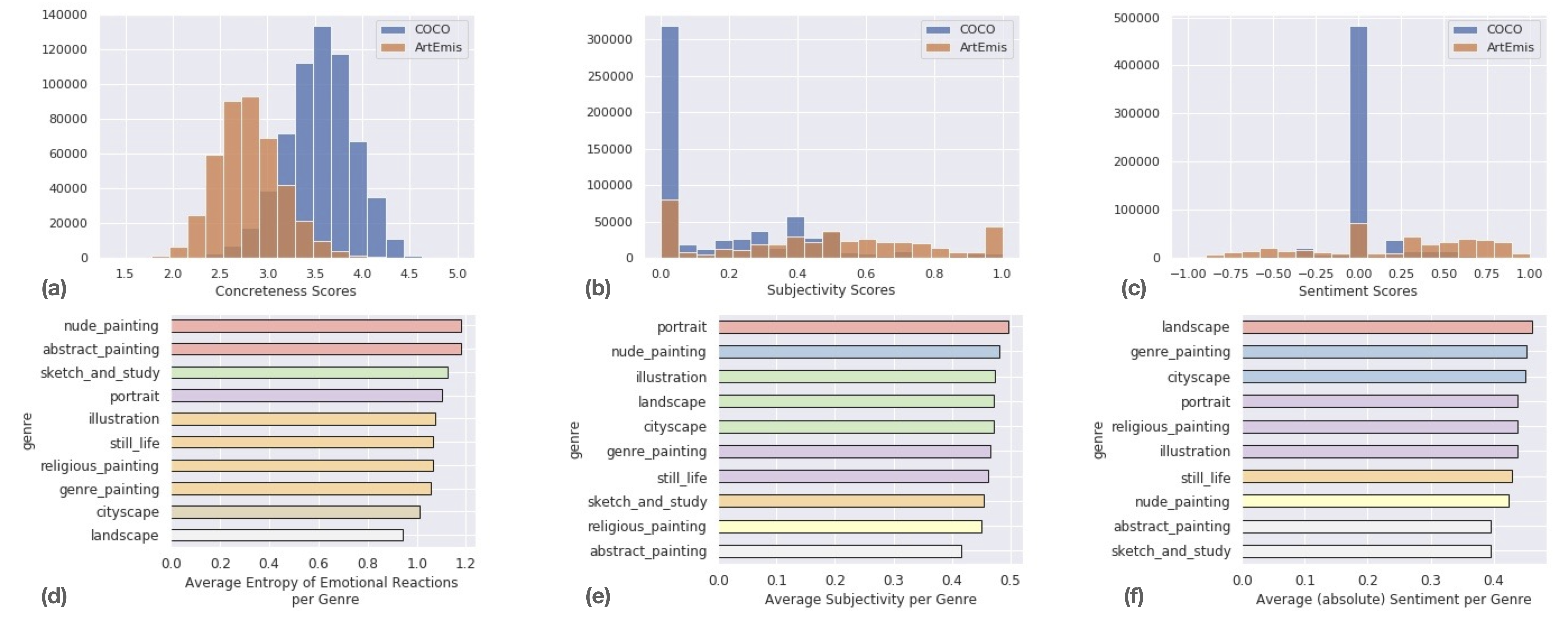}
  \caption{{\bf Key properties of {\datasetname} $\&$ Genre Oriented Analysis.}. Top-row: histograms comparing {\datasetname} to COCO-captions along the axes of (a) \textit{Concreteness}, (b) \textit{Subjectivity}, and (c) \textit{Sentiment} . {\datasetname} has significantly more abstract, subjective and sentimental language than COCO-captions. Bottom-row: (d) displays the average entropy of distributions of emotions elicited in ArtEmis across different artworks of the same art genre. (e) and (f) display averages for the \textit{Subjectivity} and \textit{Sentiment} metrics used in the top-row.}
  \label{fig:analysis_teaser}
\end{figure*}

\section{\datasetname~ dataset}
\label{sec:dataset}

The \textit{\datasetname}~ dataset is built on top of the publicly available WikiArt\footnote{\url{https://www.wikiart.org/}} dataset which contains \numimages{} carefully curated artworks from 1,119 artists (as downloaded in 2015), covering artwork created as far back as the 15\textsuperscript{th} century, to modern fine art paintings created in the 21\textsuperscript{st} century. The artworks cover 27 art-styles (abstract, baroque, cubism, impressionism, etc.) and 45 genres (cityscape, landscape, portrait, still life, etc.), constituting a very diverse set of visual stimuli~\cite{clf_of_wikiart}. In \datasetname~we annotated \textit{all} artworks of WikiArt by asking at least 5 annotators per artwork to express their dominant emotional reaction along with an utterance explaining the reason behind their response. 

Specifically, after observing an artwork, an annotator was asked first to indicate their \textit{dominant} reaction by selecting among the eight emotions mentioned in Section \ref{sec:related_work}, or a ninth option, listed as `something-else'. This latter option was put in place to allow  annotators to express emotions not explicitly listed, or to explain why they might not have had any strong emotional reaction e.g., why they felt indifferent to the shown artwork. In all cases, after the first step, the annotator was asked to provide a detailed explanation for their choice in free text that would include specific references to visual elements in the artwork. See Figures~\ref{fig:bird_annotations},\ref{fig:subjectivity-of-artemis} for examples of collected annotations and Figure~\ref{fig:data_collection_interface} for a quick overview of the used interface.

In total, we collected \textbf{\numcaptions{}} explanatory utterances and emotional responses. The resulting corpus contains \numwords{} distinct words and it includes the explanations of \numworkers{} annotators who worked in aggregate \numhours{} hours to build it. The annotators were recruited via Amazon's Mechanical Turk (AMT) services. In what follows we analyze the key characteristics of \datasetname{}, while pointing the interested reader to the Supplemental Material~\cite{artemis_supp} for further details.

\subsection{Linguistic analysis}
\label{para:linguistic_analysis}

\paragraph{Richness \& diversity.} The average length of the captions of {\datasetname} is \captionlength{} words which is significantly longer than the average length of captions of many existing captioning datasets as shown in Table~\ref{table:pos_per_captions}. In the same table, we also show results of analyzing {\datasetname} in terms of the average number of nouns, pronouns, adjectives, verbs, and adpositions. {\datasetname} has a higher occurrence per caption for each of these categories compared to many existing datasets, indicating that our annotations provide rich use of natural language in connection to the artwork and the emotion they explain. This fact becomes even more pronounced when we look at \textit{unique}, say adjectives, that are used to explain the reactions to the same artwork among different annotators (Table~\ref{table:pos_per_image}). In other words, besides being linguistically rich, the collected explanations are also highly \textit{diverse}.

\input{tables/datasets_POS_comparative_analysis_per_utterance}
\input{tables/datasets_POS_comparative_analysis_per_image}

\vspace{-5pt}
\mypara{Sentiment analysis.}
\label{para:dataset-sent-analysis}
In addition to being rich and diverse, {\datasetname} also contains language that is sentimental. We use a rule-based sentiment analyzer (VADER~\cite{VADER}) to demonstrate this point. The analyzer assigns only $16.5\%$ of {\datasetname} to the neutral sentiment, while for COCO-captions it assigns $77.4\%$. Figure~\ref{fig:analysis_teaser} ~(c)  shows the histogram of VADER's estimated valences of sentimentality for the two datasets. Absolute values closer to 0 indicate neutral sentiment. More details on this metric are in the \suppmat.


\begin{figure}[ht]
  \includegraphics[width=\linewidth]{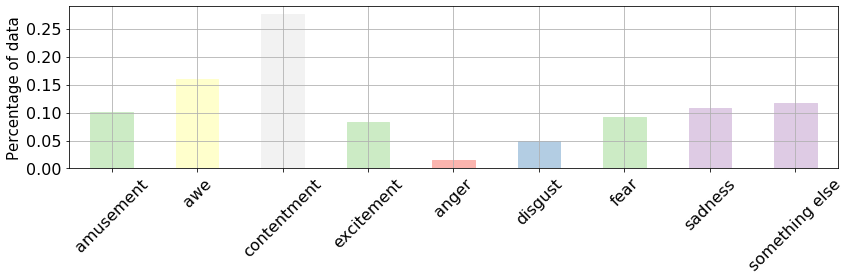}
  \vspace{-8pt}
  \caption{{\bf Histogram of emotions captured in \datasetname~}. Positive emotions occur significantly more often than negative emotions (four left-most bars contain $61.9\%$ of all responses vs. 5th-8th bars contain $26.3\%$). The annotators use a non-listed emotion (`something-else' category) $11.7\%$ of the time.}
  \label{fig:histogram_emotions_clicks}
\end{figure}

\subsection{Emotion-centric analysis.}
\label{para:emotion-analysis}

In Figure~\ref{fig:histogram_emotions_clicks} we present the histogram over the nine options that the users selected, across all collected annotations. We remark that positive emotions are chosen significantly more often than negative ones, while the ``something-else'' option was selected 11.7\%. Interestingly, 61\% of artworks have been annotated with at least one positive and one negative emotion simultaneously (this percent is 79\% if we treat something-else as a third emotion category). While this result highlights the high degree of subjectivity w.r.t.~the emotional reactions an artwork might trigger, we also note that that there is significant agreement among the annotators w.r.t.~the elicited emotions. Namely, 45.6\% (37,145) of the paintings have a strong majority among their annotators who indicated the same fine-grained emotion. 



\mypara{Idiosyncrasies of language use.} 
Here, we explore the degree to which {\datasetname} contains language that is abstract vs.~concrete, subjective vs.~objective, and estimate the extent to which annotators use similes and metaphors in their explanations. To perform this analysis we tag the collected utterances and compare them with externally curated lexicons that carry relevant meta-data. For measuring the abstractness or concreteness, we use the lexicon in Brysbaert et al.~\cite{40k_absrtact_words} which provides for 40,000 word lemmas a rating from 1 to 5 reflecting their concreteness. For instance, \textit{banana} and \textit{bagel} are maximally concrete/tangible objects, getting a score of 5, but \textit{love} and \textit{psyche} are quite abstract (with scores 2.07 and 1.34, resp.). A random word of {\datasetname} has 2.80 concreteness while a random word of COCO has 3.55 (p-val significant, see Figure~\ref{fig:analysis_teaser} (a)). In other words, {\datasetname} contains on average references to more abstract concepts. This also holds when comparing {\datasetname} to other widely adopted captioning datasets (see~\suppmat). Next, to measure the extent to which {\datasetname} makes subjective language usage, we apply the rule-based algorithm provided by TextBlob~\cite{textblob} which estimates how subjective a sentence is by providing a scalar value in $[0,1]$. E.g., \textit{`The painting is red'} is considered a maximally objective utterance (scores 1), while \textit{`The painting is nice'}, is maximally subjective (scores 0). We show the resulting distribution of these estimates in Figure~\ref{fig:analysis_teaser} (b). Last, we curated a list of lemmas that suggest the use of similes with high probability (e.g., `is like', `looks like', `reminds me of'). Such expressions appear on $20.5\%$ of our corpus and, as shown later, are also successfully adopted by our neural-speakers.

\subsection{Maturity, reasonableness \& specificity.}
\label{para:dataset-Maturity-reasonableness-specificity}
We also investigated the unique aspects of {\datasetname} by conducting three separate user studies. Specifically we aim to understand: a) what is the \textit{emotional and cognitive maturity} required by someone to express a random {\datasetname} explanation?, b) how \textit{reasonable} a human listener finds a random {\datasetname} explanation, even when they would not use it to describe their own reaction?, and last, c) to what extent the collected explanations can be used to \textit{distinguish} one artwork from another? We pose the first question to Turkers in a binary (yes/no) form, by showing to them a randomly chosen artwork and its accompanying explanation and asking them if this explanation requires emotional maturity higher than that of a typical 4-year old. The answer for 1K utterances was `yes' \textbf{76.6\%} of the time. In contrast, repeating the same experiment with the COCO dataset, the answer was positive significantly less (\textbf{34.5\%}). For the second question, we conducted an experiment driven by the question ``Do you think this is a realistic and reasonable emotional response that could have been given by someone for this image?''. Given a randomly sampled utterance, users had four options to choose, indicating the degree of response appropriateness for that artwork. We elaborate on the results in \suppmat; in summary, 97.5\% of the utterances were considered appropriate. To answer the final question, we presented Turkers with one piece of art coupled with one of its accompanying explanations, and placed it next to two random artworks, side by side and in random order. We asked Turkers to guess the `referred' piece of art in the given explanation. The Turkers succeeded in predicting the `target’ painting 94.7\% of the time in a total of 1K trials.

These findings indicate that, despite the inherent subjective nature of \datasetname, there is significant common ground in identifying a reasonable affective utterance and suggest aiming to build models that replicate such high quality captions.

%% file: tables/datasets_POS_comparative_analysis_per_utterance.tex

\begin{table}[b]
    \centering
    \resizebox{\linewidth}{!}{%
    \begin{tabular}{l@{\hspace{5mm}}c@{\hspace{4mm}}c@{\hspace{4mm}}c@{\hspace{4mm}}c@{\hspace{4mm}}c@{\hspace{4mm}}c}
        \toprule
        Dataset & Words & Nouns & Pronouns & Adjectives & Adpositions & Verbs\\
        \midrule
        \textit{ArtEmis} & \captionlength{}  & 4.0 & 0.9 & 1.6 &  1.9 & 3.0 \\
        \coco{} Captions~\cite{coco_chen2015} &  10.5 & 3.7 & 0.1 & 0.8 & 1.7 & 1.2 \\
        Conceptual Capt.~\cite{conceptual-captions}  & 9.6 & 3.8 & 0.2 & 0.9 & 1.6 & 1.1\\
        \flickr{}~\cite{young14tacl}  & 12.3 & 4.2 & 0.2 & 1.1 & 1.9 & 1.8 \\
        Google Refexp~\cite{mao16} & 8.4 & 3.0 & 0.1 & 1.0 & 1.2 & 0.8 \\
        \bottomrule
    \end{tabular}}\\[2mm]
    \caption{\small\textbf{Richness of individual captions} of ArtEmis vs. previous works. We highlight the richness of captions as units and thus show word counts averaged over \textit{individual captions}.}
    \label{table:pos_per_captions}
\end{table}

%% file: tables/datasets_POS_comparative_analysis_per_image.tex
\begin{table}[b]
    \centering
    \resizebox{\linewidth}{!}{%
    \begin{tabular}{l@{\hspace{5mm}}c@{\hspace{4mm}}c@{\hspace{4mm}}c@{\hspace{4mm}}c@{\hspace{4mm}}c}
        \toprule
        Dataset & Nouns & Pronouns & Adjectives & Adpositions & Verbs\\
        \midrule
        \textit{ArtEmis} & 17.6 (3.4) & 3.0 (0.6) & 7.7 (1.5) & 6.3 (1.2) & 12.6 (2.4) \\
        \coco{} Captions~\cite{coco_chen2015} & 10.8 (2.2) & 0.6 (0.1) & 3.3 (0.7) & 4.5 (0.9) & 4.5 (0.9) \\
        Conceptual Capt.~\cite{conceptual-captions} & 3.8 (3.8) & 0.2 (0.2) & 0.9 (0.9) & 1.6 (1.6)  & 1.1 (1.1) \\
        \flickr{}~\cite{young14tacl} & 12.9 (2.6) & 0.8 (0.2) & 4.0 (0.8) & 4.9 (1.0) & 6.4 (1.3) \\
        Google Refexp~\cite{mao16} & 7.8 (2.2) & 0.4 (0.1) & 2.8 (0.8) & 2.9 (0.8) & 2.3 (0.6) \\
        \bottomrule
    \end{tabular}}\\[2mm]
    \caption{\small\textbf{Diversity of captions per image} of ArtEmis vs. previous works. Shown are \textit{unique} word counts for various parts-of-speech averaged over \textit{individual images}. To account for discrepancies in the number of captions individual images have, we also include the correspondingly normalized averages inside parentheses.}
    \label{table:pos_per_image}
\end{table}

%% file: sections/method.tex
\section{Neural methods}
\label{sec:method}

\subsection{Auxiliary classification tasks}
\label{subsection:aux_classifiers}
Before we present the neural speakers we introduce two auxiliary \textit{classification} problems and corresponding neural-based solutions. First, we pose the problem of predicting the emotion explained with a given textual explanation of ArtEmis. This is a classical 9-way text classification problem admitting standard solutions. In our implementations we use cross-entropy-based optimization applied to an LSTM text classifier trained from scratch, and also consider fine-tuning to this task a pretrained BERT model~\cite{devlin2018bert}. 

Second, we pose the problem of predicting the expected distribution of emotional reactions that \textit{users} typically would have given an artwork. To address this problem we fine-tune an ImageNet-based~\cite{deng2009imagenet} pretrained ResNet-\resnetD{} encoder~\cite{he2016deep} by minimizing the KL-divergence between its output and the empirical user distributions of ArtEmis.
Having access to these two classifiers, which we denote as $C_{emotion|text}$ and $C_{emotion|image}$ respectively, is useful for our neural speakers as we can use them to evaluate, and also, steer, the emotional content of their output (Sections~\ref{section:evaluation} and~\ref{subsection:emotion_grounded_speaker}). Of course, these two problems have also intrinsic value and we explore them in detail in Section~\ref{sec:experimental_results}.

\subsection{Affective neural speakers}
\paragraph{Baseline with ANPs.}
In order to illustrate the importance of having an emotion-explanation-oriented dataset like ArtEmis for building affective neural speakers; we borrow ideas from previous works~\cite{sentiment_inject,SentiCap} and create a baseline speaker that does not make any (substantial) use of ArtEmis. Instead, and similar to what was done for the baseline presented in~\cite{SentiCap}, we first train a neural speaker with the COCO-caption dataset and then we inject \textit{sentiment} to its generated captions by adding to them appropriately chosen adjectives. Specifically we use the intersection of Adjective Noun Pairs (ANPs) between ArtEmis and the ANPs of \cite{SentiCap} (resulting in 1,177 ANPs, with known positive and negative sentiment) and capitalize on the $C_{emotion|image}$ to decide what sentiment we want to emulate. If the $C_{emotion|image}$ is maximized by one of the four positive emotion-classes of ArtEmis, we inject the adjective corresponding to the \textit{most frequent} (per ArtEmis) positive ANP, to a randomly selected noun of the caption. If the maximizer is negative, we use the corresponding ANP with negative sentiment; last, we resolve the something-else maximizers (\textless $10\%$) by fair coin-flipping among the two sentiments. We note that since we apply this speaker to {\datasetname} images and there is significant visual domain gap between COCO and WikiArt, we fine-tune the neural-speaker on a small-scale and separately collected (by us) dataset with \textit{objective} captions for 5,000 wikiArt paintings. We stress that this new dataset was collected following the AMT protocol used to build COCO-captions, i.e., asking only for objective (not affective) descriptions of the main objects, colors etc. present in an artwork. Examples of these annotations are in the \suppmat{}

\mypara{Basic {\datasetname} speakers.}
We experiment with two popular backbone architectures when designing neural speakers trained on {\datasetname}: the Show-Attend-Tell (SAT) approach~\cite{xu2015show}, which combines an image encoder with a word/image attentive LSTM; and the recent line of work of top-down, bottom-up meshed-memory transformers ($M^{2}$)~\cite{cornia2020meshed}, which replaces the recurrent units with transformer units and capitalizes on separately computed object-bounding-box detections (computed using Faster R-CNN~\cite{girshick2015fast}). We also include a simpler baseline that uses {\datasetname} but without training on it: for a test image we find its nearest visual neighbor in the training set (using ImageNet pre-trained ResNet-\resnetD{} features) and output a random caption associated with this neighbor.

\mypara{Emotion grounded speaker.}
\label{subsection:emotion_grounded_speaker}
We additionally tested neural speakers that make use of the emotion classifier, i.e., $C_{emotion|image}$. At training time, in addition to grounding the (SAT) neural-speaker with the visual stimulus and applying teacher forcing with the captions of {\datasetname}, we further provide at each time step a feature (extracted via a fully-connected layer) of the emotion-label chosen by the annotator for that specific explanation. This extra signal promotes the \textit{decoupling} of the emotion conveyed by the linguistic generation, from the underlying image. In other words, this speaker allows us to independently set the emotion we wish to explain for a given image. At inference time (to keep things fair) we deploy first the $C_{emotion|image}$ over the test artwork, and use the output maximizing emotion, to first ground and then sample the generation of this variant.

\mypara{Details.}
To ensure a meaningful comparison between neural-speakers, we use the same image-encoders, learning-rate schedules, LSTM hidden-dimensions, etc. across all of them. When training with {\datasetname} we use an $[85\%, 5\%, 10\%]$ train-validation-test data split and do model-selection (optimal epoch) according to the model that minimizes the negative-log-likelihood on the validation split. For the ANP baseline, we use the Karpathy splits~\cite{karpathy2015deep} to train the same (SAT) backbone network we used elsewhere. When \textit{sampling} a neural speaker, we keep the test generation with the highest log-likelihood resulting from a greedy beam-search with beam size of 5 and a soft-max temperature of 0.3. The only exception to the above (uniform) experimental protocol was made for the basic ArtEmis speaker, trained with Meshed Transformers. In this case we used the author's publicly available implementation without customization~\cite{m2_implementation}.

%% file: sections/evaluation.tex

\section{Evaluation}
\label{section:evaluation}
In this section we describe the evaluation protocol we follow to quantitatively compare our trained neural networks. First, for the auxiliary classification problems we report the average attained accuracy per method. Second, for the evaluation of the neural speakers we use three categories of metrics that assess different aspects of their quality. To measure the extent to which our generations are linguistically similar to held-out ground-truth human captions, we use various popular machine-based metrics: e.g., BLEU 1-4~\cite{BLEU}, ROUGE-L~\cite{lin2004rouge}, METEOR~\cite{denkowski:lavie:meteor-wmt:2014}. For these metrics, a higher number reflects a better agreement between the model-generated caption and at least one of the ground-truth annotator-written captions. 

We highlight that CIDEr-D~\cite{cider} which requires a generation to be semantically close to \textit{all} human-annotations of an artwork, is not a well suited metric for ArtEmis, due to the large diversity and inherent subjectivity of our dataset (see more on this on Supp. Mat). The second dimension that we use to evaluate our speakers concerns how \textit{novel} their captions are; here we report the average length of the longest common subsequence for a generation and (a subsampled version) of all training utterances. The smaller this metric is, the farther away one can assume that the generations are from the training data~\cite{fan2019strategies}. The third axis of evaluation concerns two unique properties of ArtEmis and affective explanations in particular. First, we report the percent of a speaker's productions that contain similes, i.e., generations that have lemmas like `thinking of', `looks like' etc. This percent is a proxy for how often a neural speaker chooses to utter metaphorical-like content. Secondly, by tapping on the $C_{emotion|text}$, we can compute which emotion is most likely explained by the generated utterance; this estimate allows us to measure the extent to which the deduced emotion is `aligned' with some ground-truth. Specifically, for test artworks where the emotion annotations form a strong majority, we define the \textit{emotional-alignment} as the percent of the grounded generations where the $\argmax(C_{emotion|generation})$ agrees to the emotion made by the majority. 

The above metrics are algorithmic, i.e., they do not involve direct \textit{human judgement}, which is regarded as the golden standard for quality assessment~\cite{cui2018learning,kilickaya2016re} of synthetic captions. The discrepancy between machine and human-based evaluations can be exacerbated in a dataset with subjective and affective components like ArtEmis. To address this, we evaluate our two strongest (per machine metrics) speaker variants via user studies that imitate a Turing test; i.e., they assess the extent to which the synthetic captions can be `confused' as being made by humans.

\begin{figure*}[htbp]
\centering
\includegraphics[width=\textwidth]{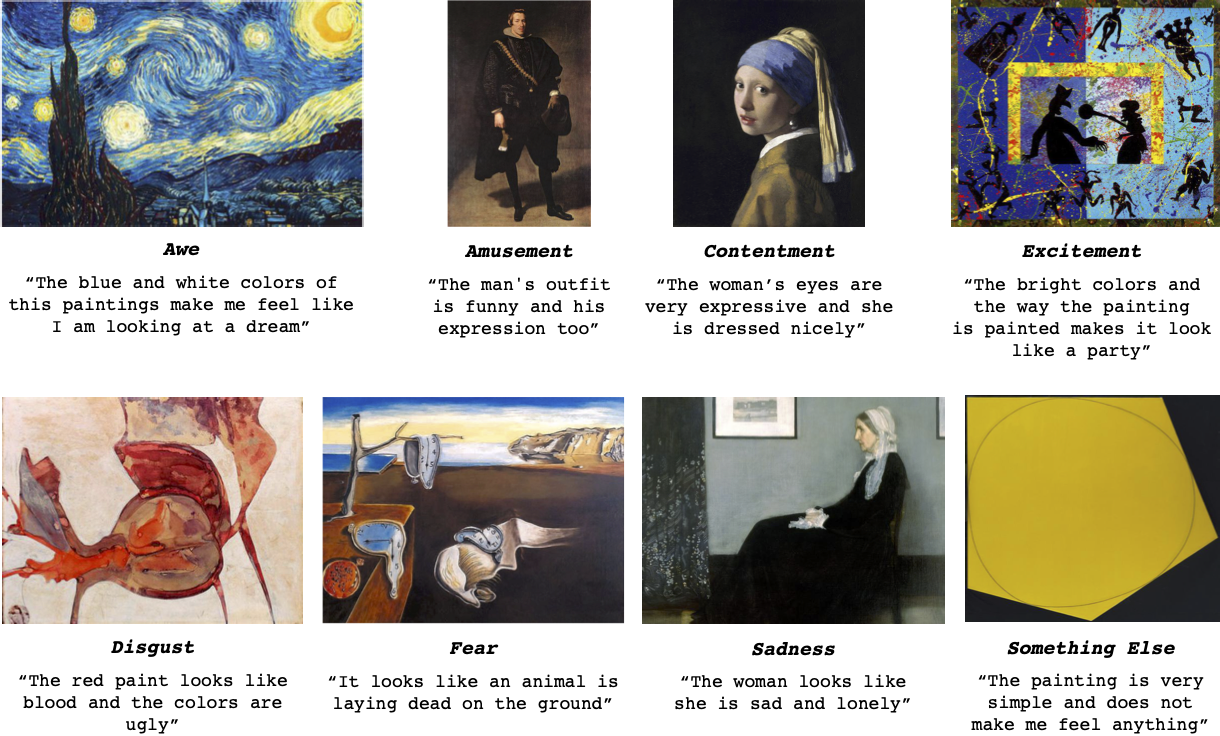}
\caption{\textbf{Examples of neural speaker productions on \textit{unseen} artworks.} The produced explanations reflect a variety of dominant emotional-responses (shown above each utterance in bold font). The top row shows examples where the deduced grounding emotion was positive; the bottom row shows three examples where the deduced emotion was negative and an example from the something-else category. Remarkably, the neural speaker can produce pragmatic explanations that include \textbf{visual analogies}: \textit{looks like blood, like a dead animal}, and \textbf{nuanced} explanations of affect: \textit{sad and lonely, expressive eyes}.}
\label{fig:neural_productions}
\vspace{-5pt}
\end{figure*}

%% file: sections/results.tex

\section{Experimental results}
\label{sec:experimental_results}

\paragraph{Estimating emotion from text or images alone.} 
We found experimentally that predicting the fine-grained emotion explained in ArtEmis data is a difficult task (see examples where both humans and machines fail in Table~\ref{table:hard_examples_to_guess_emotion}). An initial AMT study concluded that users where able to infer the exact emotion from text alone 53.0\% accurately (in 1K trials). Due to this low score, we decided to make a study with experts (authors of this paper). We attained slightly better accuracy (60.3\%  on a sample of 250 utterances). Interestingly, the neural networks of Section~\ref{subsection:aux_classifiers} attained $63.1\%$ and $65.7\%$ (LSTM, BERT respectively) on the entire test split used by the neural-speakers (\numtestutterances{} utterances). Crucially, both humans and neural-nets failed gracefully in their predictions and most confusion happened among subclasses of the same, positive or negative category (we include confusion matrices in the \suppmat{}). For instance, w.r.t. binary labels of positive vs. negative emotion sentiment (ignoring the something-else annotations), the experts, the LSTM and the BERT model, guess correctly $85.9\%$, $87.4\%$, $91.0\%$ of the time. This is despite being trained, or asked in the human studies, to solve the fine-grained 9 way problem.
\vspace{5pt}
\input{tables/hard_examples_to_guess_emotion}

Since we train our image classifiers to predict a distribution of emotions, we select the maximizer of their output and compare it with the `dominant' emotion of the (8,160) test images for which the emotion distribution is unimodal with a mode covering more than $50\%$ of the mass ($38.5\%$ of the split). The attained accuracy for this sub-population is $60.0\%$. We note that the training (and test) data are highly unbalanced, following the emotion-label distribution indicated by the histogram of Figure~ \ref{fig:histogram_emotions_clicks}. As such, losses addressing long-tail, imbalanced classification problems (e.g.,\cite{lin2017focal}) could be useful in this setup.

\mypara{Neural speakers.}
In Table~\ref{table:speaker_metrics} we report the machine-induced metrics described in Section~\ref{section:evaluation}. First, we observe that on metrics that measure the linguistic similarity to the held-out utterances (BLEU, METEOR, etc.) the speakers fare noticeably worse as compared to how the same architectures fare (modulo secondary-order details) when trained and tested with objective datasets like COCO-captions; e.g., BLEU-1 with SOTA ~\cite{cornia2020meshed} is 82.0. This is expected given the analysis of Section~\ref{sec:dataset} that shows how ArtEmis is a more diverse and subjective dataset. Second, there is a noticeable difference in all metrics in favor of the three models trained with ArtEmis (denoted as Basic or Grounded) against the simpler baselines that do not. This implies that we cannot simply reproduce ArtEmis with ANP injection on objective data. It further demonstrates how even among similar images the annotations can be widely different, limiting the Nearest-Neighbor (NN) performance. Third, on the emotion-alignment metric (denoted as Emo-Align) the emotion-grounded variant fares significantly better than its non-grounded version. This variant also produces a more appropriate percentage of similes by staying closest to the ground-truth's percentage of $~20.5$.

Qualitative results of the emotion-grounded speaker are shown in Figure~\ref{fig:neural_productions}. More examples, including typical failure cases and generations from other variants, are provided in the project's website\footnote{\url{\webpage}} and the \suppmat{} As seen in Figure~\ref{fig:neural_productions} a well-trained speaker creates sophisticated explanations that can incorporate nuanced emotional understanding and analogy making.

\mypara{Turing test.} For our last experiment, we performed a user study taking the form of a Turing Test deployed in AMT. First, we use a neural-speaker to make one explanation for a test artwork and couple it with a randomly chosen ground-truth for the same stimulus. Next, we show to a user the two utterances in text, along with the artwork, and ask them to make a multiple choice among 4 options. These were to indicate either that one utterance was more likely than the other as being made by a human explaining their emotional reaction; or, to indicate that both (or none) were likely made by a human. We deploy this experiment with 500 artworks, and repeat it separately for the basic and the emotion-grounded (SAT) speakers. Encouragingly, \textbf{50.3\%} of the time the users signaled that the utterances of the emotion-grounded speaker were on-par with the human groundtruth (20.6\%, were selected as the more human-like of the pair, and 29.7\% scored a tie). Furthermore, the emotion-grounded variant achieved  significantly better results than the basic speaker, which surpassed or tied to the human annotations 40\% of the time (16.3\%  with a win and and 23.7\% as a tie). To explain this differential, we hypothesize that grounding with the \textit{most likely} emotion of the $C_{emotion|image}$ helped the better-performing variant to create more common and thus on average more fitting explanations which were easier to pass as being made by a human.

\vspace{-13pt}
\mypara{Limitations.} While these results are encouraging, we also remark that the quality of even the best neural speakers is very far from human ground truth, in terms of diversity, accuracy and creativity of the synthesized utterances. Thus, significant research is necessary to bridge the gap between human and synthetic emotional neural speakers. We hope that {\datasetname} will enable such future work and pave the way towards a deeper and nuanced emotional image understanding.
\input{tables/speaker_metrics}

%% file: tables/hard_examples_to_guess_emotion.tex
\begin{table}[h]
    \centering
    \resizebox{\linewidth}{!}{%
    \begin{tabular}{c@{\hspace{3mm}}c@{\hspace{3mm}}c@{\hspace{3mm}}}
        \toprule
        ArtEmis Utterance & Guess & GT\\
        \midrule
        ``The scene reminds me of a perfect & \multirow{2}{*}{Contentment (H)} & \multirow{2}{*}{Awe}  \\ 
        summer day." & & \\ \hline
        ``This looks like me when I don't want to & \multirow{2}{*}{Something-Else (M)} & \multirow{2}{*}{Amusement} \\ 
        get out of bed on Monday morning." & & \\ \hline
        ``A proper mourning scene, and the & \multirow{2}{*}{Sadness (H)} & \multirow{2}{*}{Contentment} \\ 
        mood is fitting." & & \\
        \bottomrule
    \end{tabular}}\\[2mm]
    \caption{\small\textbf{Examples showcasing why fine-grained emotion-deduction from text is hard.} The first two examples' interpretation depends highly on personal experience (first \& middle row). The third example uses language that is emotionally subtle. (H): human-guess, (M): neural-net guess, GT: ground-truth.}
    \label{table:hard_examples_to_guess_emotion}
\end{table}

%% file: tables/speaker_metrics.tex

\begin{table}[t]
    \centering
    \resizebox{\linewidth}{!}{%
    \begin{tabular}{l@{\hspace{5mm}}c@{\hspace{4mm}}c@{\hspace{4mm}}c@{\hspace{4mm}}c@{\hspace{4mm}}c@{\hspace{4mm}}c}
        \toprule
         metric       &       NN  &       ANP &    Basic($M^2$) & Basic(SAT) & Grounded(SAT)  \\
        \midrule
            BLEU-1    &     0.346 &     0.386 & 0.484 & \textbf{0.505} &  \textbf{0.505} \\
            BLEU-2    &     0.119 &     0.124 & 0.251 & \textbf{0.254} &  0.252 \\
            BLEU-3    &     0.055 &     0.059 & \textbf{0.137} & 0.132 &  0.130\\
            BLEU-4    &     0.035 &     0.039 & \textbf{0.088} & 0.081 &  0.080 \\
            METEOR    &     0.100 &     0.087 & 0.137 &  \textbf{0.139}  &  0.137 \\
            ROUGE-L   &     0.208 &     0.204 & 0.280 &  \textbf{0.295}  &  0.293 \\
            max-LCS   &     8.296 &     \textbf{5.646} & 7.868 &  7.128  & 7.346 \\
            mean-LCS  &     1.909 &     \textbf{1.238} & 1.630 &  1.824  & 1.846 \\
            Emo-Align &     0.326 &     0.451 & 0.385 &  0.400  &  \textbf{0.522} \\
            Similes-percent &     \textbf{0.197} &     0.000 & 0.675 &  0.452 &   0.356 \\
        \bottomrule
    \end{tabular}}\\[2mm]
    \caption{\small\textbf{Neural speaker machine-based evaluations}. NN: Nearest Neighbor baseline, ANP: baseline-with-injected sentiments, $M^2$: Meshed Transformer, SAT: Show-Attend-Tell. The Basic models use for grounding only the underlying image, while the Grounded variant also inputs an emotion-label.
    For details see Section~\ref{sec:method}.}.
    \label{table:speaker_metrics}
    \vspace{-13pt}
\end{table}


%% file: sections/conclusion.tex

\section{Conclusion}
\label{sec:conclusion}

Human cognition has a strong affective component that has been relatively undeveloped in AI systems. Language that explains emotions generated at the sight of a visual stimulus gives us a way to analyze how image content is related to affect, enabling learning that can lead to agents emulating human emotional responses through data-driven approaches. In this paper, we take the first step in this direction through: (1) the release of the {\datasetname} dataset that focuses on linguistic explanations for affective responses triggered by visual artworks with abundant emotion-provoking content; and (2) a demonstration of neural speakers that can express emotions and provide associated explanations. The ability to deal computationally with images' emotional attributes opens an exciting new direction in human-computer communication and interaction.

%% file: sections/acknowledgments.tex
\paragraph{\noindent {\bf Acknowledgements.}} This work is funded by 
a Vannevar Bush Faculty Fellowship, a KAUST BAS/1/1685-01-01, a CRG-2017-3426, the ERC Starting Grant No. 758800 (EXPROTEA) and the ANR AI Chair AIGRETTE, and gifts from the Adobe, Amazon AWS, Autodesk, and Snap corporations. The authors wish to thank Fei Xia and Jan Dombrowski for their help with the AMT instruction design and Nikos Gkanatsios for several fruitful discussions. The authors also want to emphasize their gratitude to all the hard working Amazon Mechanical Turkers without whom this work would not be possible.

%% file: main.bbl
\begin{thebibliography}{10}\itemsep=-1pt

\bibitem{achlioptas2020referit3d}
Panos Achlioptas, Ahmed Abdelreheem, Fei Xia, Mohamed Elhoseiny, and Leonidas
  Guibas.
\newblock Referit3d: Neural listeners for fine-grained 3d object identification
  in real-world scenes.
\newblock {\em European Conference on Computer Vision (ECCV)}, 2020.

\bibitem{achlioptas2019shapeglot}
Panos Achlioptas, Judy Fan, Robert~XD Hawkins, Noah~D Goodman, and Leonidas~J
  Guibas.
\newblock {ShapeGlot}: Learning language for shape differentiation.
\newblock In {\em International Conference on Computer Vision (ICCV)}, 2019.

\bibitem{artemis_supp}
Panos Achlioptas, Maks Ovsjanikov, Kilichbek Haydarov, Mohamed Elhoseiny, and
  Leonidas Guibas.
\newblock {\em Supplementary Material for ArtEmis: Affective Language for
  Visual Art}, (accessed January 1st, 2021).
\newblock Available at \url{\webpage/materials/artemis_supplemental.pdf},
  version 1.0.

\bibitem{adolphs2017should}
Ralph Adolphs.
\newblock How should neuroscience study emotions? by distinguishing emotion
  states, concepts, and experiences.
\newblock {\em Social cognitive and affective neuroscience}, 2017.

\bibitem{barrett2006solving}
Lisa~Feldman Barrett.
\newblock Solving the emotion paradox: Categorization and the experience of
  emotion.
\newblock {\em Personality and social psychology review}, 2006.

\bibitem{barrett2017emotions}
Lisa~Feldman Barrett.
\newblock {\em How emotions are made: The secret life of the brain}.
\newblock Houghton Mifflin Harcourt, 2017.

\bibitem{barrett2017theory}
Lisa~Feldman Barrett.
\newblock The theory of constructed emotion: an active inference account of
  interoception and categorization.
\newblock {\em Social cognitive and affective neuroscience}, 2017.

\bibitem{barrett2007mice}
Lisa~Feldman Barrett, Kristen~A Lindquist, Eliza Bliss-Moreau, Seth Duncan,
  Maria Gendron, Jennifer Mize, and Lauren Brennan.
\newblock Of mice and men: Natural kinds of emotions in the mammalian brain? a
  response to panksepp and izard.
\newblock {\em Perspectives on Psychological Science}, 2007.

\bibitem{barrett2019historical}
Lisa~Feldman Barrett and Ajay~B Satpute.
\newblock Historical pitfalls and new directions in the neuroscience of
  emotion.
\newblock {\em Neuroscience letters}, 693:9--18, 2019.

\bibitem{vector_model_of_affect}
M.~M. Bradley, M.~K. Greenwald, M.C. Petry, and P.~J. Lang.
\newblock Remembering pictures: Pleasure and arousal in memory.
\newblock {\em Journal of Experimental Psychology: Learning, Memory, and
  Cognition}, 1992.

\bibitem{IAPS}
Margaret~M. Bradley and Peter~J. Lang.
\newblock The international affective picture system (iaps) in the study of
  emotion and attention.
\newblock {\em Series in affective science. Handbook of emotion elicitation and
  assessment}, 2007.

\bibitem{40k_absrtact_words}
Marc Brysbaert, Amy Warriner, and Victor Kuperman.
\newblock Concreteness ratings for 40 thousand generally known english word
  lemmas.
\newblock {\em Behavior Research Methods}, 2014.

\bibitem{caba2015activitynet}
Fabian Caba~Heilbron, Victor Escorcia, Bernard Ghanem, and Juan Carlos~Niebles.
\newblock Activitynet: A large-scale video benchmark for human activity
  understanding.
\newblock In {\em Conference on Computer Vision and Pattern Recognition
  (CVPR)}, 2015.

\bibitem{coco_chen2015}
Xinlei Chen, Hao Fang, Tsung-Yi Lin, Ramakrishna Vedantam, Saurabh Gupta, Piotr
  Dollar, and Lawrence~C. Zitnick.
\newblock Microsoft {COCO} captions: Data collection and evaluation server.
\newblock {\em CoRR}, abs/1504.00325, 2015.

\bibitem{cornia2020meshed}
Marcella Cornia, Matteo Stefanini, Lorenzo Baraldi, and Rita Cucchiara.
\newblock Meshed-memory transformer for image captioning.
\newblock In {\em Conference on Computer Vision and Pattern Recognition
  (CVPR)}, 2020.

\bibitem{m2_implementation}
Marcella Cornia, Matteo Stefanini, Lorenzo Baraldi, and Rita Cucchiara.
\newblock {\em Meshed-Memory Transformer for Image Captioning}, (accessed
  September 1st, 2020).
\newblock Available at
  \url{https://github.com/aimagelab/meshed-memory-transformer}.

\bibitem{cui2018learning}
Yin Cui, Guandao Yang, Andreas Veit, Xun Huang, and Serge Belongie.
\newblock Learning to evaluate image captioning.
\newblock In {\em Conference on Computer Vision and Pattern Recognition
  (CVPR)}, 2018.

\bibitem{deng2009imagenet}
Jia Deng, Wei Dong, Richard Socher, Li-Jia Li, Kai Li, and Li Fei-Fei.
\newblock Imagenet: A large-scale hierarchical image database.
\newblock In {\em Conference on Computer Vision and Pattern Recognition
  (CVPR)}, 2009.

\bibitem{denkowski:lavie:meteor-wmt:2014}
Michael Denkowski and Alon Lavie.
\newblock Meteor universal: Language specific translation evaluation for any
  target language.
\newblock In {\em Proceedings of the EACL 2014 Workshop on Statistical Machine
  Translation}, 2014.

\bibitem{devlin2018bert}
Jacob Devlin, Ming-Wei Chang, Kenton Lee, and Kristina Toutanova.
\newblock Bert: Pre-training of deep bidirectional transformers for language
  understanding.
\newblock {\em CoRR}, abs/1810.04805, 2018.

\bibitem{happiness_1}
Ed Diener, Christie~Napa Scollon, and Richard~E Lucas.
\newblock The evolving concept of subjective well-being: The multifaceted
  nature of happiness.
\newblock {\em Advances in Cell Aging \& Gerontology}, 2003.

\bibitem{ekman_emotions}
Paul Ekman.
\newblock An argument for basic emotions.
\newblock {\em Cognition and Emotion}, 1992.

\bibitem{fan2019strategies}
Angela Fan, Mike Lewis, and Yann Dauphin.
\newblock Strategies for structuring story generation.
\newblock {\em CoRR}, abs/1902.01109, 2019.

\bibitem{sem_art}
Noa Garcia and George Vogiatzis.
\newblock How to read paintings: Semantic art understanding with multi-modal
  retrieval.
\newblock {\em CoRR}, abs/1810.09617, 2018.

\bibitem{girshick2015fast}
Ross Girshick.
\newblock Fast r-cnn.
\newblock In {\em Conference on Computer Vision and Pattern Recognition
  (CVPR)}, 2015.

\bibitem{he2016deep}
Kaiming He, Xiangyu Zhang, Shaoqing Ren, and Jian Sun.
\newblock Deep residual learning for image recognition.
\newblock In {\em Conference on Computer Vision and Pattern Recognition
  (CVPR)}, 2016.

\bibitem{lstm}
Sepp Hochreiter and J{\"u}rgen Schmidhuber.
\newblock Long short-term memory.
\newblock {\em Neural Computation}, 1997.

\bibitem{VADER}
C.J. Hutto and Eric~E. Gilbert.
\newblock Vader: A parsimonious rule-based model for sentiment analysis of
  social media text. eighth international conference on weblogs and social
  media.
\newblock {\em ICWSM}, 2014.

\bibitem{karpathy2015deep}
Andrej Karpathy and Li Fei-Fei.
\newblock Deep visual-semantic alignments for generating image descriptions.
\newblock In {\em Conference on Computer Vision and Pattern Recognition
  (CVPR)}, 2015.

\bibitem{prose_for_painting}
Prerna Kashyap, Samrat Phatale, and Iddo Drori.
\newblock Prose for a painting.
\newblock {\em CoRR}, abs/1910.03634, 2019.

\bibitem{Kazemzadeh}
Sahar Kazemzadeh, Vicente Ordonez, Mark Matten, and L.~Tamara Berg.
\newblock Referitgame: Referring to objects in photographs of natural scenes.
\newblock {\em Conference on Empirical Methods in Natural Language Processing
  (EMNLP)}, 2014.

\bibitem{kilickaya2016re}
Mert Kilickaya, Aykut Erdem, Nazli Ikizler-Cinbis, and Erkut Erdem.
\newblock Re-evaluating automatic metrics for image captioning.
\newblock {\em CoRR}, abs/1612.07600, 2016.

\bibitem{emotion_clf_valence_arousal}
H. {Kim}, Y. {Kim}, S.~J. {Kim}, and I. {Lee}.
\newblock Building emotional machines: Recognizing image emotions through deep
  neural networks.
\newblock {\em IEEE Transactions on Multimedia}, 2018.

\bibitem{VG_Krishna_2017}
Ranjay Krishna, Yuke Zhu, Oliver Groth, Justin Johnson, Kenji Hata, Joshua
  Kravitz, Stephanie Chen, Yannis Kalantidis, Li-Jia Li, David~A. Shamma, and
  et al.
\newblock Visual genome: Connecting language and vision using crowdsourced
  dense image annotations.
\newblock {\em International Journal of Computer Vision}, 2017.

\bibitem{lin2004rouge}
Chin-Yew Lin.
\newblock Rouge: A package for automatic evaluation of summaries.
\newblock In {\em Text summarization branches out}, 2004.

\bibitem{lin2017focal}
Tsung-Yi Lin, Priya Goyal, Ross Girshick, Kaiming He, and Piotr Doll{\'a}r.
\newblock Focal loss for dense object detection.
\newblock In {\em International Conference on Computer Vision (ICCV)}, 2017.

\bibitem{lindquist2015role}
Kristen~A Lindquist, Jennifer~K MacCormack, and Holly Shablack.
\newblock The role of language in emotion: Predictions from psychological
  constructionism.
\newblock {\em Frontiers in psychology}, 2015.

\bibitem{textblob}
Steven Loria.
\newblock {\em TextBlob}, (accessed November 16, 2020).
\newblock Available at \url{https://textblob.readthedocs.io/en/dev/}.

\bibitem{baby_talk}
Jiasen Lu, Jianwei Yang, Dhruv Batra, and Devi Parikh.
\newblock Neural baby talk.
\newblock {\em Conference on Computer Vision and Pattern Recognition (CVPR)},
  2018.

\bibitem{img_clf_art}
Jana Machajdik and Allan Hanbury.
\newblock Affective image classification using features inspired by psychology
  and art theory.
\newblock In {\em ACM International Conference on Multimedia}, 2010.

\bibitem{mao16}
Junhua Mao, Jonathan Huang, Alexander Toshev, Oana Camburu, Alan Yuille, and
  Murphy Kevin.
\newblock Generation and comprehension of unambiguous object descriptions.
\newblock {\em CoRR}, abs/1511.02283, 2016.

\bibitem{Mikels_2005}
Joseph~A. Mikels, Barbara~L. Fredrickson, Gregory~R. Larkin, Casey~M Lindberg,
  Sam~J. Maglio, and Patricia~A Reuter-Lorenz.
\newblock Emotional category data on images from the international affective
  picture system.
\newblock {\em Behavior research methods}, 2005.

\bibitem{monroe_colors}
Will Monroe, Robert~X.D. Hawkins, Noah~D. Goodman, and Christopher Potts.
\newblock Colors in context: A pragmatic neural model for grounded language
  understanding.
\newblock {\em CoRR}, abs/1703.10186, 2017.

\bibitem{nagaraja16}
K.~Varun Nagaraja, I.~Vlad Morariu, and Davis~S. Larry.
\newblock Modeling context between objects for referring expression
  understanding.
\newblock {\em European Conference on Computer Vision (ECCV)}, 2016.

\bibitem{OrtonyBook}
Andrew Ortony, Gerald~L. Clore, and Collins Allan.
\newblock {\em The Cognitive Structure of Emotions}.
\newblock Cambridge University Press, 1988.

\bibitem{BLEU}
Kishore Papineni, Salim Roukos, Todd Ward, and Wei-Jing Zhu.
\newblock Bleu: A method for automatic evaluation of machine translation.
\newblock In {\em Proceedings of the 40th Annual Meeting on Association for
  Computational Linguistics}, ACL '02, 2002.

\bibitem{SentiCap}
Alexander~Mathews Patrick, Lexing Xie, and Xuming He.
\newblock Senticap: Generating image descriptions with sentiments.
\newblock {\em AAAI}, 2016.

\bibitem{pont2020connecting}
Jordi Pont-Tuset, Jasper Uijlings, Soravit Changpinyo, Radu Soricut, and
  Vittorio Ferrari.
\newblock Connecting vision and language with localized narratives.
\newblock In {\em European Conference on Computer Vision (ECCV)}, 2020.

\bibitem{emotion_clf}
You Quanzeng, Luo Jiebo, Jin Hailin, and Yang Jianchao.
\newblock Building a large scale dataset for image emotion recognition: The
  fine print and the benchmark.
\newblock {\em CoRR}, abs/1605.02677, 2016.

\bibitem{unsup_sentiment_cell}
Alec Radford, Rafal Jozefowicz, and Ilya Sutskever.
\newblock Learning to generate reviews and discovering sentiment.
\newblock {\em CoRR}, abs/1704.01444, 2017.

\bibitem{compare_dim_models}
David Rubin and Jennifer Talarico.
\newblock A comparison of dimensional models of emotion: Evidence from
  emotions, prototypical events, autobiographical memories, and words.
\newblock {\em Memory (Hove, England)}, 2009.

\bibitem{circumplex_model}
James Russell.
\newblock A circumplex model of affect.
\newblock {\em Journal of Personality and Social Psychology}, 1980.

\bibitem{clf_of_wikiart}
Babak Saleh and Ahmed Elgammal.
\newblock Large-scale classification of fine-art paintings: Learning the right
  metric on the right feature.
\newblock {\em CoRR}, abs/1505.00855, 2015.

\bibitem{shackman2019emotional}
Alexander~J Shackman and Tor~D Wager.
\newblock The emotional brain: Fundamental questions and strategies for future
  research.
\newblock {\em Neuroscience letters}, 693:68, 2019.

\bibitem{conceptual-captions}
Piyush Sharma, Nan Ding, Sebastian Goodman, and Radu Soricut.
\newblock Conceptual captions: A cleaned, hypernymed, image alt-text dataset
  for automatic image captioning.
\newblock In {\em Annual Meeting of the Association for Computational
  Linguistics (ACL)}, 2018.

\bibitem{tolstoy-art}
Leo Tolstoy.
\newblock {\em What is Art?}
\newblock London: Penguin, 1995 [1897].

\bibitem{unsup_context_aware}
Ramakrishna Vedanta, Samy Bengio, Kevin Murphy, Devi Parikh, and Gal Chechik.
\newblock Context-aware captions from context-agnostic supervision.
\newblock {\em CoRR}, abs/1701.02870, 2017.

\bibitem{cider}
R. {Vedantam}, C.~L. {Zitnick}, and D. {Parikh}.
\newblock Cider: Consensus-based image description evaluation.
\newblock In {\em Conference on Computer Vision and Pattern Recognition
  (CVPR)}, 2015.

\bibitem{show-tell}
Oriol Vinyals, Alexander Toshev, Samy Bengio, and Dumitru Erhan.
\newblock Show and tell: A neural image caption generator.
\newblock {\em CoRR}, abs/1411.4555, 2015.

\bibitem{Wilber2017BAMTB}
J.~Michael Wilber, Chen Fang, H. Jin, Aaron Hertzmann, J. Collomosse, and
  J.~Serge Belongie.
\newblock Bam! the behance artistic media dataset for recognition beyond
  photography.
\newblock {\em International Conference on Computer Vision (ICCV)}, 2017.

\bibitem{teacher_forcing}
Ronald~J. Williams and David Zipser.
\newblock A learning algorithm for continually running fully recurrent neural
  networks.
\newblock {\em Neural Computation}, 1989.

\bibitem{xu2015show}
Kelvin Xu, Jimmy Ba, Ryan Kiros, Kyunghyun Cho, Aaron Courville, Ruslan
  Salakhudinov, Rich Zemel, and Yoshua Bengio.
\newblock Show, attend and tell: Neural image caption generation with visual
  attention.
\newblock In {\em International Conference on Machine Learning (ICML)}, 2015.

\bibitem{Yanulevskaya-emotions}
Victoria Yanulevskaya, Jan Gemert, Katharina Roth, Ann-Katrin Schild, Nicu
  Sebe, and Jan-Mark Geusebroek.
\newblock Emotional valence categorization using holistic image features.
\newblock In {\em IEEE International Conference on Image Processing}, 2008.

\bibitem{sentiment_inject}
Quanzeng You, Hailin Jin, and Luo Jiebo.
\newblock Image captioning at will: A versatile scheme for effectively
  injecting sentiments into image descriptions.
\newblock {\em CoRR}, abs/1801.10121, 2018.

\bibitem{young14tacl}
Peter Young, Alice Lai, Micah Hodosh, and Julia Hockenmaier.
\newblock From image descriptions to visual denotations: New similarity metrics
  for semantic inference over event descriptions.
\newblock {\em Transactions of the Association for Computational Linguistics},
  2014.

\bibitem{mattnet}
Licheng Yu, Zhe Lin, Xiaohui Shen, Yangm Jimei, Xin Lu, Mohit Bansal, and
  L.~Tamara Berg.
\newblock Mattnet: Modular attention network for referring expression
  comprehension.
\newblock {\em Conference on Computer Vision and Pattern Recognition (CVPR)},
  2018.

\bibitem{licheng_16}
Licheng Yu, Patrick Poirson, Shan Yang, C.~Alexander Berg, and L.~Tamara Berg.
\newblock Modeling context in referring expressions.
\newblock {\em European Conference on Computer Vision (ECCV)}, 2016.

\bibitem{Zhao2014art}
Sicheng Zhao, Yue Gao, Xiaolei Jiang, Hongxun Yao, Tat-Seng Chua, and Xiaoshuai
  Sun.
\newblock Exploring principles-of-art features for image emotion recognition.
\newblock In {\em ACM International Conference on Multimedia}, 2014.

\end{thebibliography}
